\documentclass[final,twocolumn,12pt,number]{elsarticle}

\usepackage[margin=1in]{geometry}

\usepackage{amsmath}
\usepackage{amssymb}

\usepackage{graphicx}
\usepackage{booktabs}
\usepackage{array}
\usepackage{longtable}
\usepackage{multirow}
\usepackage{tabularx}
\usepackage{dblfloatfix}

\usepackage[hidelinks]{hyperref}

\usepackage{threeparttable}
\usepackage{makecell}
\usepackage{ragged2e}
\usepackage{placeins}

\biboptions{sort&compress}
\journal{Decision Support Systems}

\begin{document}

% \begin{graphicalabstract}
% \centering
% \includegraphics[width=\linewidth]{figs/Graphical Abstract_Mufti et al.,-01.png}
% \end{graphicalabstract}

% =========================
% FRONTMATTER (ANONYMIZED)
% =========================
\begin{frontmatter}

\title{%
A Rolling-Window Framework for Churn Prediction and Behavioral Driver Identification
}

% IMPORTANT (Double anonymized review):
% Do NOT put author names, affiliations, acknowledgements, funding, or COI here.
% Those go in a separate TITLE PAGE file.

\author[1]{Muhammad Jawad Mufti}
% \cormark[1]
\ead{g202392310@kfupm.edu.sa}
% \credit{Writing -- original draft \& editing; Methodology; Formal analysis; Validation; Investigation.}

\author[1]{Omar Hammad}
% \credit{Conceptualization; Resources; Methodology; Supervision; Project Administration; Writing -- review \& editing.}

\affiliation[1]{organization={Information and Computer Science Department, King Fahd University of Petroleum and Minerals},
                city={Dhahran},
                postcode={31261},
                country={Saudi Arabia}}

\author[2]{Haitham Saleh}
% \credit{Conceptualization; Methodology; Project Administration; Writing -- review \& editing.}

\affiliation[2]{organization={Interdisciplinary Research Center for Smart Mobility and Logistics (IRC-SML), King Fahd University of Petroleum and Minerals},
                city={Dhahran},
                postcode={31261},
                country={Saudi Arabia}}

\author[3]{Muqaddas Gull}
% \credit{Investigation; Writing -- review \& editing.}

\affiliation[3]{organization={SDAIA--KFUPM Joint Research Center for Artificial Intelligence, King Fahd University of Petroleum and Minerals},
                city={Dhahran},
                postcode={31261},
                country={Saudi Arabia}}

\cortext[cor1]{Corresponding author.}

% NOTE:
% DSS encourages Highlights and Graphical Abstract, but they should be submitted
% as separate files in the submission system (recommended).
% So we do NOT embed them here.

\begin{abstract}
Customer churn prediction is a central task in customer analytics, particularly in non-contractual, pay-per-use service environments where disengagement is not explicitly observed and must be inferred from behavioral inactivity. Existing churn prediction approaches often rely on simplified temporal assumptions or single-point representations of customer behavior, which limit their ability to support continuous risk assessment, interpretability, and realistic deployment over time. This study proposes a temporally explicit churn prediction framework that models customer behavior usingrolling behavioral windows, enabling repeated and instance-level churn risk estimationas customer activity evolves. Customer behavior is summarized within a fixed 30-days observation window, followed by a 30-days future churn evaluation window, ensuring a clear temporal separation between behavioral evidence and churn outcomes. The framework integrates feature-based and sequence-based learning approaches within a unified temporal design. The proposed approach is evaluated on a large-scale, real-world dataset from a non-contractual service platform. Empirical results demonstrates trong and stable predictive performance, with accuracy reaching 87.6\% and ROC-AUC of 0.94 for the feature-based model, while the sequence-based model achieves recall as high as 96.1\% by capturing temporal disengagement patterns. Evaluation on future unseen data confirms meaningful robustness under temporal shift, with accuracy remaining above 83\% and ROC-AUC exceeding 0.91 without model retraining. Overall, the findings highlight that carefully designed temporal framing, rather than model complexity alone, is critical for achieving robust, interpretable, and deployment-ready churn prediction, providing a practical foundation for churn-oriented decision support in dynamic service environments.
\end{abstract}


\end{frontmatter}

% =========================
% MAIN TEXT (NUMBERED SECTIONS)
% =========================

\section{Introduction} \label{sec:introduction}

Customer churn represents a persistent challenge across service-oriented and usage-based industries, where retaining existing customers is often more cost-effective than acquiring new ones. As competitive pressures increase and customer switching costs decline, organizations increasingly rely on predictive analytics to anticipate customer disengagement and to support retention-oriented strategies. In this context, customer churn prediction models are commonly used to estimate the likelihood that a customer will cease active usage of a service based on historical behavioral patterns, transactional activity, and usage signals \citep{DeCaigny2018,Verbeke2012}. Accurate churn prediction enables organizations to prioritize at-risk customers and to allocate limited retention resources more effectively, making churn modeling a central task in contemporary customer analytics \citep{Neslin2006,Ascarza2018}.

In applied churn analytics, predictive accuracy alone is insufficient if model outputs cannot be meaningfully interpreted. For churn prediction to be practically useful, analysts must be able to identify the behavioral and transactional factors that contribute to attrition risk, as such insights support diagnosis and subsequent intervention \citep{Coussement2017}. In parallel, prior research has shown that the evaluation of churn models should extend beyond predictive accuracy to reflect their operational usefulness, particularly in contexts where model outputs inform retention actions \citep{Verbraken2013}. Consequently, churn prediction models are increasingly expected to balance predictive performance with explanatory capability, as failing to satisfy either objective can limit their practical relevance.

Despite the extensive body of research on customer churn prediction, several methodological limitations continue to constrain the development of decision-oriented and deployment-ready churn prediction systems. A large proportion of existing studies formulate churn prediction as a static classification task, in which each customer is represented by a single data instance and assigned a single churn label, even when rich transactional or usage data are available \citep{DeCaigny2020,Lalwani2022,Geiler2022,DeCaigny2024}. Although recent studies have incorporated temporal information through sequence learning, panel data, or sliding-window designs, these approaches often operate at fixed or sequence-level representations and do not support continuous, rolling churn risk reassessment for the same customer over time \citep{Mena2024,Ahlstrand2025,Bugajev2025}. Furthermore, many studies lack an explicit and consistent temporal problem formulation, with observation windows and churn horizons either undefined or implicitly assumed, thereby limiting interpretability and real-world applicability \citep{Vo2021,Krishna2024,Chajia2024}. Even when raw behavioral data are available, they are frequently aggregated into static or coarse-grained summaries, such as lifetime RFM features or quarterly usage statistics, which can obscure short-term behavioral dynamics that may precede churn \citep{Sanchez2024,Asfe2025}. Finally, the majority of churn prediction frameworks are designed for contractual business models with explicit churn events, while comparatively fewer studies address non-contractual, pay-per-use service contexts in which churn must be inferred from sustained inactivity patterns \citep{Zaghloul2025,Bugajev2025}. Collectively, these limitations highlight the need for churn prediction frameworks that explicitly define temporal windows, preserve fine-grained behavioral dynamics, and support rolling, instance-level churn risk assessment in non-contractual service settings.

To address these gaps, this study proposes a temporally explicit churn prediction framework designed for non-contractual, pay-per-use service environments. The proposed approach models customer behavior using rolling behavioral windows that advance over time, enabling repeated churn risk assessment for the same customer as new activity data becomes available. Customer engagement is summarized within explicitly defined observation windows, while churn is operationalized through a subsequent evaluation window, ensuring a clear temporal separation between behavioral evidence and churn outcomes. By constructing multiple time-indexed instances per customer, the framework captures short-term behavioral dynamics that are often obscured by static or lifetime aggregation. The approach integrates feature-based and sequence-based learning techniques within a unified temporal design and incorporates interpretable model outputs to support decision-making in proactive customer retention scenarios.

The remainder of this paper is organized as follows. Section~\ref{sec:related-work} summarizes existing research on customer churn prediction, with an emphasis on temporal modeling approaches and decision-support considerations. Section~\ref{sec:methodology} describes the proposed methodology, including the temporal windowing strategy and modeling framework. Section~\ref{sec:results} presents the empirical results. Section~\ref{sec:discussion} discusses the findings in the context of decision support and deployment considerations. Finally, Section~\ref{sec:conclusion} concludes the paper and outlines directions for future research.

\section{Related Work} \label{sec:related-work}
Customer churn prediction has been extensively studied across a wide range of service and subscription-based domains, with the primary objective of identifying customers who are likely to discontinue their relationship with a firm. Early and widely adopted approaches frame churn prediction as a static binary classification task, where each customer is represented by a single feature vector summarizing historical behavior. Such formulations are commonly used in benchmark-driven studies that compare traditional machine learning classifiers, including logistic regression, random forests, gradient boosting, and ensemble methods, often reporting strong predictive performance on cross-sectional datasets \citep{Lalwani2022,Geiler2022,Krishna2024}. These studies demonstrate that supervised learning models can effectively discriminate between churners and non-churners when churn is treated as a single-point outcome, but they typically rely on static representations that abstract away temporal dynamics.

To enhance predictive performance and feature representation, several studies incorporate richer data sources and advanced modeling techniques. Unstructured data, such as textual customer interactions and call logs, have been integrated alongside structured behavioral features, yielding measurable gains in predictive accuracy \citep{DeCaigny2020, Vo2021}. More recent work has explored representation learning through embeddings and deep learning architectures, including neural networks, meta-modeling frameworks, and large language model embeddings, intending to capture complex nonlinear relationships in customer behavior \citep{Chajia2024, Asfe2025}. Hybrid and ensemble-based approaches further combine multiple modeling paradigms to balance predictive performance and interpretability, demonstrating consistent improvements over single-model baselines across multiple datasets \citep{DeCaigny2024, Zaghloul2025}.

Beyond static modeling, an increasing body of research acknowledges the temporal nature of customer behavior and seeks to incorporate time into churn prediction. One stream of work adopts panel-based or sequence-aware formulations, where customer behavior is represented through time-varying features or ordered sequences. For example, time-varying RFM measures and recurrent neural networks have been shown to improve churn prediction by capturing longitudinal patterns in customer engagement \citep{Mena2024}. Other studies employ sliding or window-based designs to summarize recent behavioral history, particularly in settings where churn is inferred from inactivity rather than explicit cancellation events \citep{Bugajev2025, Ahlstrand2025}. These approaches demonstrate the value of temporal information, yet they differ substantially in how observation windows, prediction horizons, and instance generation are defined.

Despite these advances, the literature exhibits notable inconsistencies in how the churn prediction task is temporally formulated. In many studies, the length of the observation window and the definition of the churn horizon are either implicit or loosely specified, making it difficult to compare results across studies or to interpret model outputs in an operational context \citep{Vo2021, Lalwani2022}. Even when inactivity-based churn definitions are used, churn is often inferred retrospectively without a clearly articulated future prediction window, particularly in non-contractual or usage-based service settings \citep{Zaghloul2025, Asfe2025}. As a result, while prior work provides strong evidence that both advanced modeling techniques and temporal information can enhance churn prediction, there remains limited consensus on how to formulate churn prediction as a temporally explicit and repeatedly evaluated task that aligns with evolving customer behavior.

\section{Methodology} \label{sec:methodology}

\begin{figure*}[!t]
\centering
\includegraphics[width=\textwidth]{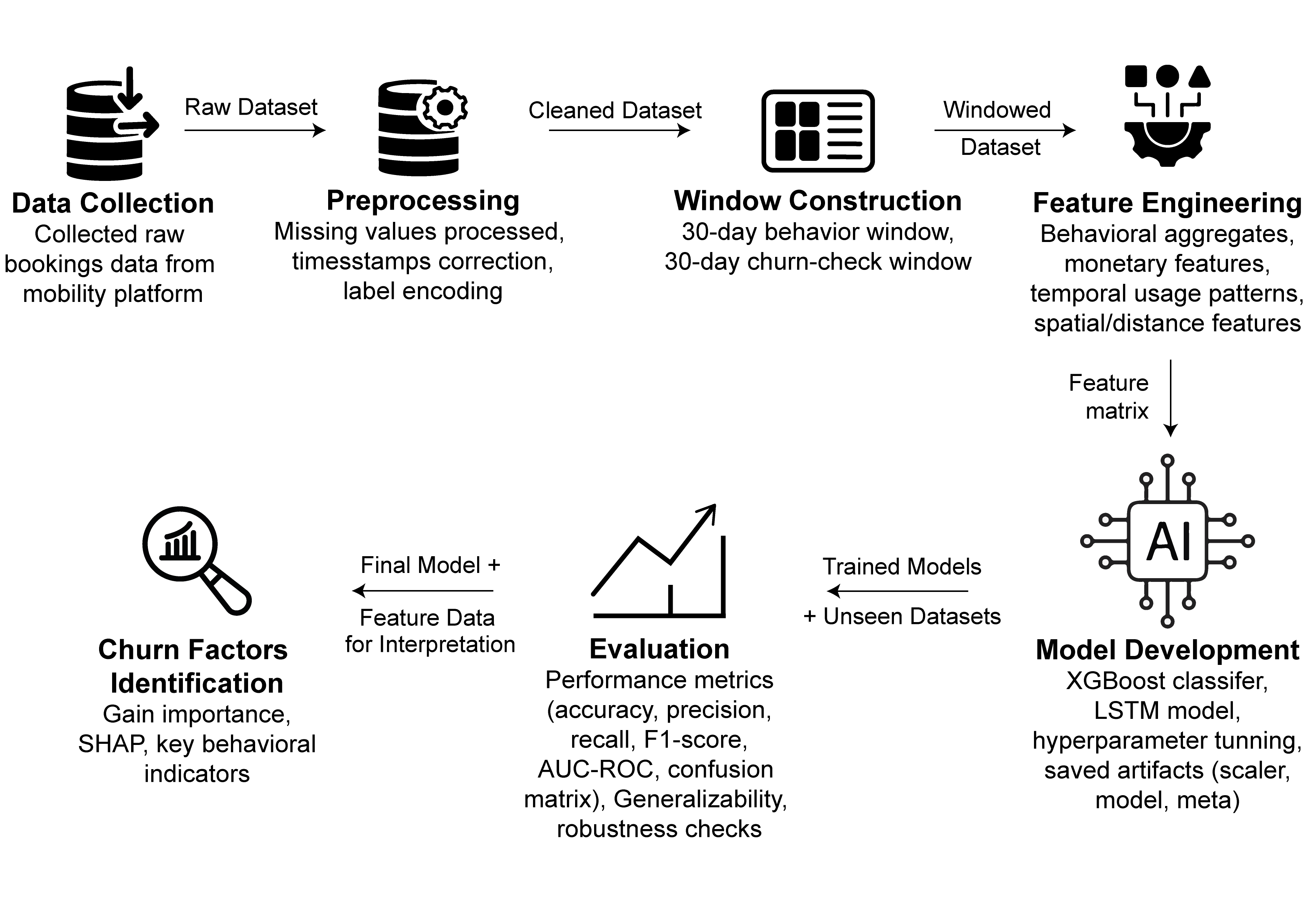}
\caption{Overview of the proposed churn prediction methodology. Raw event-level booking data are preprocessed and organized into rolling behavioral observation windows followed by a future churn evaluation horizon, ensuring temporal separation between inputs and outcomes. The resulting representations are used for feature-based and sequence-based modeling, followed by time-aware evaluation and interpretability analysis.}
\label{fig:methodology_workflow}
\end{figure*}

This section presents the proposed churn prediction framework, which is designed to model short-term behavioral dynamics in non-contractual, pay-per-use service environments. The methodology combines a rolling-window instance construction strategy with feature-based and sequence-based learning approaches, enabling repeated churn risk assessment for the same customer over time. Figure~\ref{fig:methodology_workflow} summarizes the end-to-end workflow. The overall workflow consists of data preprocessing, temporal windowing, feature engineering, model development, evaluation, and interpretability analysis.

\subsection{Problem Definition}
\label{sec:problem-definition}

Let $i \in \{1, \dots, N\}$ denote a customer and let $t$ represent a discrete time index measured in days. Customer behavior is observed through a sequence of time-stamped service interactions. The objective is to predict whether a customer will churn within a predefined future period based on their recent behavioral history.

Churn prediction is formulated as a binary classification task at the instance level. For each customer $i$ and reference time $t$, an input instance is constructed using a behavioral observation window of length $W_b$, followed by a churn evaluation window of length $W_c$. The churn label $y_{i,t}$ is defined as:

\begin{equation}
\label{eq:churn_label}
y_{i,t} =
\begin{cases}
1, & \begin{aligned} & \text{if customer } i \text{ exhibits no} \\ & \text{qualifying activity during the} \\ & \text{interval } [t+1,\, t+W_c], \end{aligned} \\[4pt] 0, & \text{otherwise}.
\end{cases}
\end{equation}

This formulation ensures a clear temporal separation between behavioral evidence and churn outcomes and aligns the prediction task with prospective churn risk assessment.

\subsection{Dataset Description and Preparation}
\label{sec:dataset-preparation}

The empirical evaluation is conducted using a real-world dataset obtained from a commercial on-demand car wash service operating under a non-contractual, pay-per-use model. The dataset consists of time-stamped service booking records collected over a continuous four-month period from October 2024 to January 2025. Each record corresponds to an individual service interaction and includes operational, transactional, temporal, and spatial attributes.

In its raw form, the dataset comprises 401{,}164 event-level booking records generated by 101{,}765 unique customers. User engagement is highly heterogeneous and temporally sparse, with a median of two bookings and one active day per user. More than half of customers (57.4\%) generate two or fewer bookings, and approximately 60\% exhibit activity spanning no more than seven days, reflecting short-lived and bursty usage patterns typical of non-contractual service environments. These characteristics motivate the use of an inactivity-based churn formulation with a fixed future evaluation window to ensure fair churn labeling for sparsely active users.

Due to the highly granular and irregular nature of the raw interaction logs, event-level data are not used directly for model training. Instead, customer activity is summarized through a rolling-window instance construction procedure (described in Section~\ref{sec:windowing}), which transforms booking events into a time-indexed panel of behavioral instances and allows each customer to contribute multiple observation windows over time.

Prior to window construction, the raw booking data undergo a structured preparation process. Preprocessing includes correction of inconsistent data types, normalization of numerical attributes, and encoding of categorical variables such as service type, booking status, and payment method, ensuring consistency and comparability across behavioral windows. Missing values are predominantly structural and arise mainly in monetary and duration-related attributes associated with unsuccessful bookings. As these attributes are meaningful only for completed services, they are computed exclusively from successful bookings (status = 2). For observation windows without successful activity, a dedicated binary indicator captures the absence of completed bookings, and all success-dependent attributes are set to zero, preserving the semantic distinction between inactivity and missing data.

Spatial attributes are derived by computing geographic distances between customers and service providers using the Haversine formula and discretizing these distances into ordinal categories based on empirical quantiles to capture spatial engagement patterns while maintaining robustness to outliers. All preprocessing steps are applied consistently across training, validation, and test datasets to prevent information leakage.

After applying rolling-window aggregation and churn labeling, the event-level data are transformed into 280{,}756 behavioral instances. Each instance represents a customer summarized over a single observation window and is described by a fixed-length feature vector of 40 engineered attributes. The resulting instance-level dataset exhibits a moderate class imbalance, with approximately 60.5\% non-churn instances and 39.5\% churn instances, reflecting realistic disengagement dynamics in non-contractual, on-demand service settings. Although the data originate from a car wash service context, both the dataset characteristics and the inactivity-based churn formulation are representative of a broader class of booking-based service platforms.

\subsection{Rolling Window Instance Construction}
\label{sec:windowing}

\begin{figure}[t]
    \centering
    \includegraphics[width=\columnwidth]{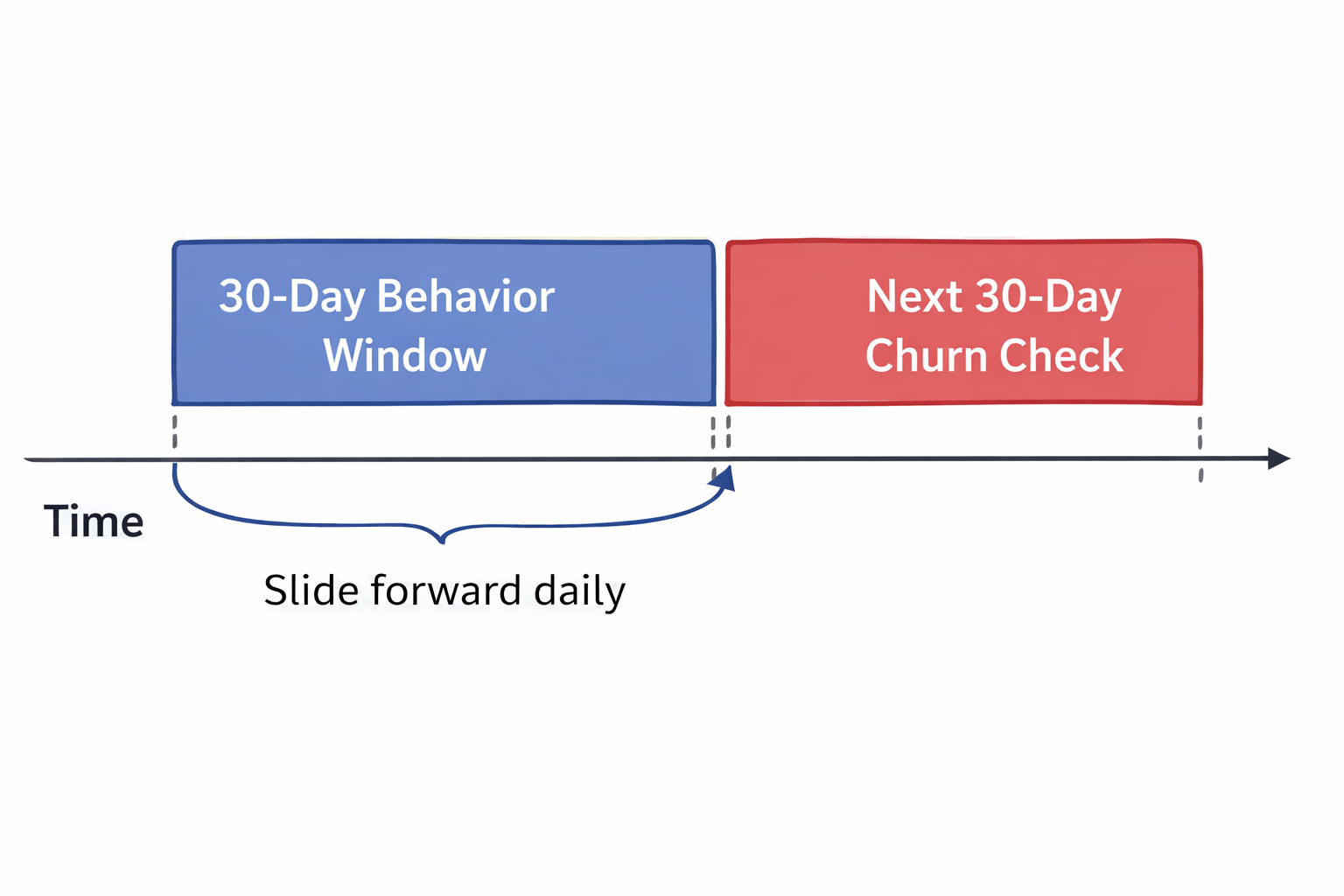}
    \caption{Rolling-window formulation}
    \label{fig:rolling_window}
\end{figure}

To capture evolving customer behavior, a rolling-window strategy is employed. For each customer $i$, a behavioral observation window of length $W_b = 30$ days is constructed, summarizing activity over the interval $[t - W_b + 1, t]$. This window is followed by a churn evaluation window of length $W_c = 30$ days, spanning $[t + 1, t + W_c]$. As illustrated in Figure~\ref{fig:rolling_window}, windows advance with a daily stride, enabling continuous reassessment of churn risk as new behavioral data become available.

Although rolling windows advance with a daily stride, customers are not treated as independent or new entities across windows. Instead, each window represents a distinct temporal state of the same customer. To avoid redundant instances, consecutive windows that yield identical behavioral summaries are collapsed, and a new instance is generated only when a behavioral change occurs, operationalized by the arrival of a new booking event within the observation window. Importantly, churn labels are assigned exclusively based on future activity, ensuring that each instance is given an equal and fair opportunity to exhibit sustained inactivity before being labeled as churn and that no information from the evaluation window leaks into feature construction.

\subsection{Feature Engineering}
\label{sec:feature-engineering}

From each behavioral observation window, a comprehensive set of behavioral and operational features is constructed to characterize customer engagement, service usage, and short-term behavioral evolution. The feature design follows a structured taxonomy to ensure interpretability, reproducibility, and alignment with the temporal churn formulation.

The first group of features captures \emph{booking activity and outcomes}, summarizing the volume of user interactions and the distribution of booking results within the observation window. These features reflect overall engagement intensity as well as the relative prevalence of successful, cancelled, and failed bookings, thereby providing insight into service reliability and user commitment.

The second group represents \emph{monetary behavior}, describing spending patterns and

\begin{table*}[!t]
\small
\caption{Categorization of engineered behavioral and operational features extracted from each observation window.}
\label{tab:feature_engineering}
\centering
\renewcommand{\arraystretch}{1.15}
\setlength{\tabcolsep}{5pt}

\begin{tabularx}{\textwidth}{p{4.2cm} X}
\toprule
\textbf{Category} & \textbf{Features} \\
\midrule
\textbf{Booking Activity} &
total\_bookings, completed\_bookings, cancelled\_bookings, failed\_bookings,
completion\_rate, cancellation\_rate, failure\_rate, only\_one\_booking,
active\_days, last\_booking\_status \\
\midrule
\textbf{Monetary Behavior} &
avg\_paid, avg\_total, avg\_discount, avg\_wallet\_usage, avg\_wash\_duration,
has\_successful\_booking \\
\midrule
\textbf{Spatial \& Service Characteristics} &
pct\_dist\_very\_near, pct\_dist\_near, pct\_dist\_mid,
pct\_dist\_mid\_far, pct\_dist\_far, services\_type\_most\_frequent \\
\midrule
\textbf{Temporal Context \& Recency} &
days\_since\_last\_booking, avg\_booking\_hour, weekend\_success\_rate,
first\_booking\_month, first\_booking\_day, last\_booking\_month, last\_booking\_day \\
\midrule
\textbf{Categorical Preferences} &
payment\_type\_most\_frequent, promocodes\_type\_most\_frequent,
discount\_type\_most\_frequent, sources\_most\_frequent,
is\_cash\_on\_delivery\_most\_frequent, use\_wallet\_most\_frequent,
booking\_type\_most\_frequent \\
\midrule
\textbf{Trend-Based Indicators} &
trend\_total\_bookings, trend\_avg\_paid, trend\_avg\_wallet\_usage,
trend\_completion\_rate \\
\bottomrule
\end{tabularx}
\end{table*}

payment-related characteristics derived exclusively from successful bookings. These features capture average payment amounts, discounts, wallet usage, and service duration, enabling the model to distinguish between high- and low-value engagement patterns.

A third category focuses on \emph{spatial behavior and service characteristics}. These features summarize the geographic dispersion of successful bookings across distance-based buckets and capture dominant service-type preferences, thereby reflecting accessibility, convenience, and service heterogeneity.

The fourth group encodes \emph{temporal context and recency}, including indicators of activity recency, temporal usage patterns, and calendar-related attributes. These features describe when customers interact with the service and how recently engagement has occurred, which are critical signals in non-contractual churn settings.

The fifth category consists of \emph{categorical preference indicators}, defined as the most frequently observed values of selected categorical attributes within the observation window. These features capture stable behavioral preferences related to payment methods, promotions, booking sources, and service configurations.

Finally, \emph{trend-based features} are computed to capture short-term behavioral dynamics within the observation window. These features quantify directional changes in engagement, spending, and success rates by estimating linear trends over time-ordered observations, thereby enabling early detection of disengagement patterns that may precede churn.

All success-dependent attributes are computed exclusively from completed bookings. For observation windows with no successful activity, a dedicated indicator variable captures the absence of successful engagement, and corresponding success-dependent attributes are set to zero to preserve the semantic distinction between inactivity and missing data. Prior to model training, all numerical features are standardized. Table~\ref{tab:feature_engineering} provides a complete overview of the engineered features grouped by category.

\subsection{Modeling Approaches}
\label{sec:modeling-approaches}

Two complementary modeling paradigms are employed to evaluate the effectiveness of aggregated versus sequential representations of customer behavior.

\subsubsection{Feature-Based Modeling}
\label{sec:xgboost}

A gradient boosting decision tree model is employed to learn churn patterns from aggregated feature vectors derived from each behavioral observation window. Gradient boosting is particularly well suited for structured and tabular data, as it effectively captures nonlinear relationships and complex feature interactions while maintaining strong generalization capability \cite{Chen2016}. The model outputs probabilistic churn scores by optimizing a binary logistic objective, making it suitable for risk-based churn prediction in decision support settings.

By operating on window-level behavioral summaries, the feature-based model serves as a strong and interpretable baseline for assessing the predictive value of short-term behavioral aggregation. Its ability to model heterogeneous engagement signals and handle mixed feature types provides a robust point of comparison against sequence-based approaches that explicitly model temporal dependencies.

\subsubsection{Sequence-Based Modeling}
\label{sec:lstm}

To explicitly model temporal dependencies in customer behavior, a Long Short-Term Memory (LSTM) network is employed on sequences of behavioral observation windows. LSTM networks are designed to address the vanishing gradient problem inherent in recurrent neural networks and are well suited for capturing long-range dependencies in sequential data \cite{Hochreiter1997}. Each input sequence represents an ordered series of window-level behavioral summaries corresponding to a customer’s recent activity history.

The sequence-based model complements the feature-based approach by learning temporal patterns that may not be fully captured by static aggregation alone, such as gradual disengagement or evolving usage trends. By leveraging sequential representations of customer behavior, the LSTM enables a comparative evaluation of feature-based versus sequence-based learning under the same rolling-window formulation.

\subsection{Model Configuration}
\label{sec:model-config}
Table~\ref{tab:model_config} summarizes the final configurations used for the feature-based and sequence-based models. For XGBoost, hyperparameters controlling tree complexity, learning rate, and subsampling were selected through grid-based search on the validation data to balance predictive performance and generalization. Subsampling and column sampling were employed to reduce overfitting in the high-dimensional behavioral feature space.

For the sequence-based model, the LSTM architecture and training parameters were selected using validation performance as the primary criterion. Early stopping based on validation loss was applied to prevent overfitting and to ensure stable convergence. These configurations were fixed after model development and used consistently across both test evaluation and temporal generalizability experiments.

\begin{table*}[t]
\centering
\caption{Final model configurations used in the experiments}
\label{tab:model_config}
\begin{tabular}{lll}
\hline
\textbf{Model} & \textbf{Parameter} & \textbf{Value} \\
\hline
XGBoost 
& Number of estimators & 400 \\
& Maximum tree depth & 8 \\
& Learning rate & 0.05 \\
& Subsample ratio & 0.8 \\
& Column subsample by tree & 0.8 \\
& Objective function & Binary logistic \\
\hline
LSTM 
& Number of LSTM layers & 2 \\
& Hidden units per layer & 32 \\
& Sequence length (time steps) & 10 \\
& Padding strategy & Left padding with training-feature medians \\
& Masking strategy & Packed sequences using true lengths \\
& Dropout rate & 0.4 \\
& Output activation & Sigmoid \\
& Loss function & Binary cross-entropy \\
& Optimizer & Adam \\
& Learning rate & $5 \times 10^{-4}$ \\
& Weight decay (L2) & $1 \times 10^{-4}$ \\
& Batch size & 32 \\
& Maximum epochs & 100 \\
& Early stopping patience & 7 \\
\hline
\end{tabular}
\end{table*}

\subsection{Evaluation Protocol}
\label{sec:evaluation}

Model performance is evaluated using Accuracy, Precision, Recall, F1-score, and the Area Under the Receiver Operating Characteristic Curve (ROC-AUC). ROC-AUC is particularly suitable for churn prediction tasks as it provides a threshold-independent measure of discriminative performance and is robust to class imbalance \cite{Fawcett2006}.

To reflect realistic deployment conditions, model development is conducted exclusively on historical data collected during 2024 using a time-aware data split rather than random sampling. Time-aware evaluation has been shown to provide more reliable performance estimates for predictive models applied to temporal and behavioral data, as random splits can introduce information leakage and lead to overly optimistic results \cite{Bergmeir2018}. Models are trained, validated, and tested on the 2024 data, after which the final trained models are fixed and saved.

To assess temporal generalizability under realistic deployment conditions, the pre-trained models are subsequently applied to a temporally held-out dataset representing future customer behavior from June-July 2025. This future dataset is processed using the same preprocessing, rolling-window construction, feature engineering, and churn labeling pipeline as the training data, and no retraining or parameter updates are performed. This out-of-time evaluation provides an explicit assessment of model robustness to temporal shifts in customer behavior and distributional change.

\subsection{Churn Factor Identification Methods}
\label{sec:explainability}

To support interpretation of model predictions, both global and local explainability techniques are applied. Gain-based feature importance is used to assess the relative contribution of predictors in the feature-based model based on their influence on ensemble split decisions \cite{Breiman2001}. In addition, Shapley Additive Explanations (SHAP) are employed to provide instance-level feature attributions for individual predictions \cite{Lundberg2017}.

\section{Results}\label{sec:results}
This section presents the empirical results of the proposed churn prediction framework. Model performance is evaluated using the metrics and evaluation protocol defined in Section~\ref{sec:evaluation}, including both rolling-window test data and a temporally held-out dataset representing future customer bookings. Results are reported for the feature-based XGBoost model and the sequence-based LSTM model, followed by an assessment of their temporal generalizability and an analysis of the factors contributing to churn predictions.

\begin{table*}[t]
\centering
\caption{Predictive performance of XGBoost and LSTM models on rolling-window data}
\label{tab:model_performance}
\begin{tabular}{lcccccc}
\hline
\textbf{Model} & \textbf{Dataset} & \textbf{Accuracy} & \textbf{Precision} & \textbf{Recall} & \textbf{F1-score} & \textbf{ROC-AUC} \\
\hline
XGBoost & Training & 89.5\% & 90.0\% & 90.0\% & 90.0\% & -- \\
XGBoost & Test     & 87.6\% & 88.0\% & 88.0\% & 88.0\% & 0.941 \\
\hline
LSTM    & Training & 90.67\% & -- & -- & -- & -- \\
LSTM    & Validation & 90.16\% & -- & -- & -- & -- \\
LSTM    & Test     & 90.1\% & -- & 96.1\% & -- & 0.940 \\
\hline
\end{tabular}
\end{table*}

\begin{figure*}[t]
\centering
\includegraphics[width=0.45\textwidth]{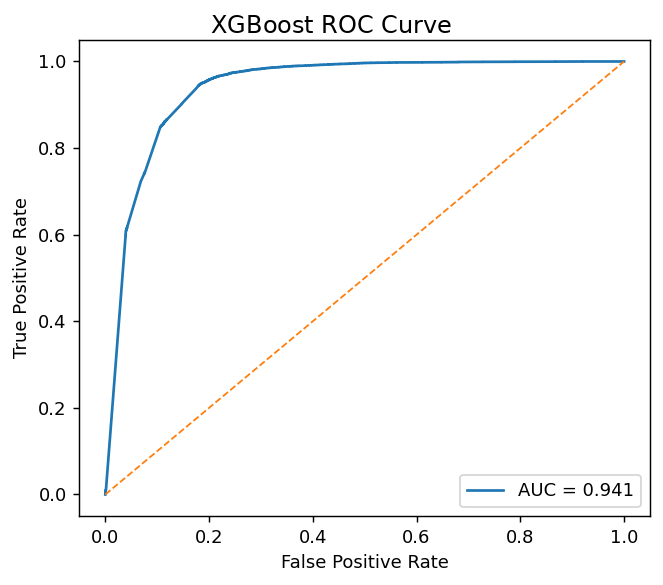}
\hfill
\includegraphics[width=0.45\textwidth]{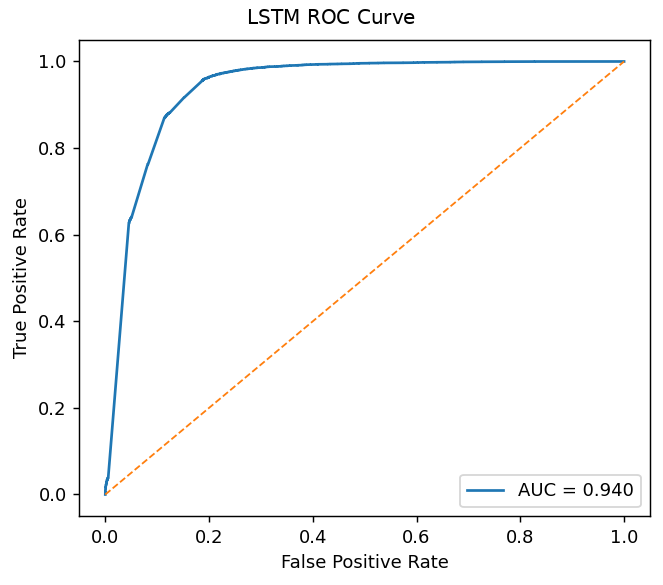}
\caption{ROC curves for (a) XGBoost and (b) LSTM models on rolling-window test data.}
\label{fig:roc_comparison}
\end{figure*}

\begin{figure*}[t]
\centering
\includegraphics[width=0.45\textwidth]{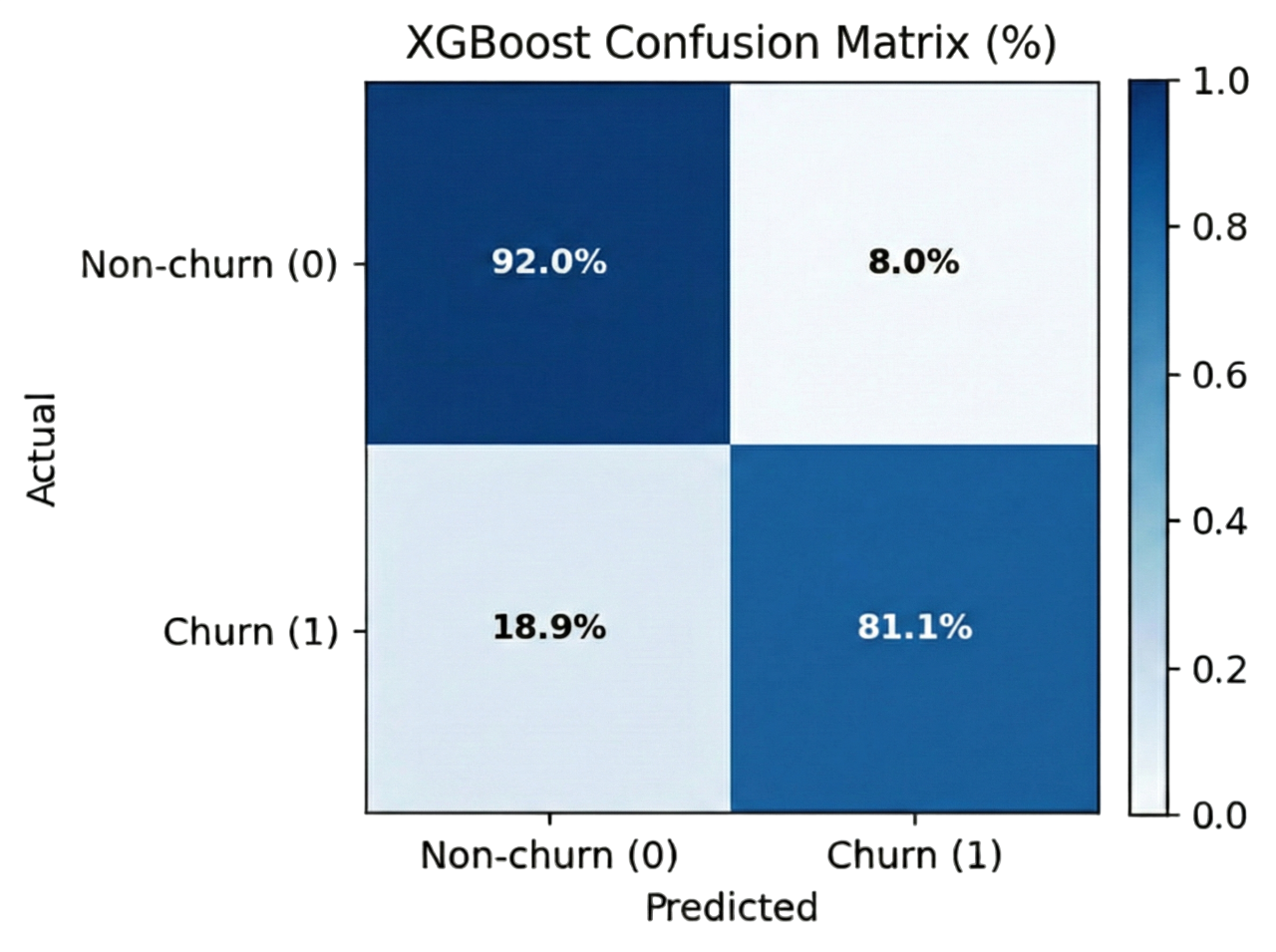}
\hfill
\includegraphics[width=0.45\textwidth]{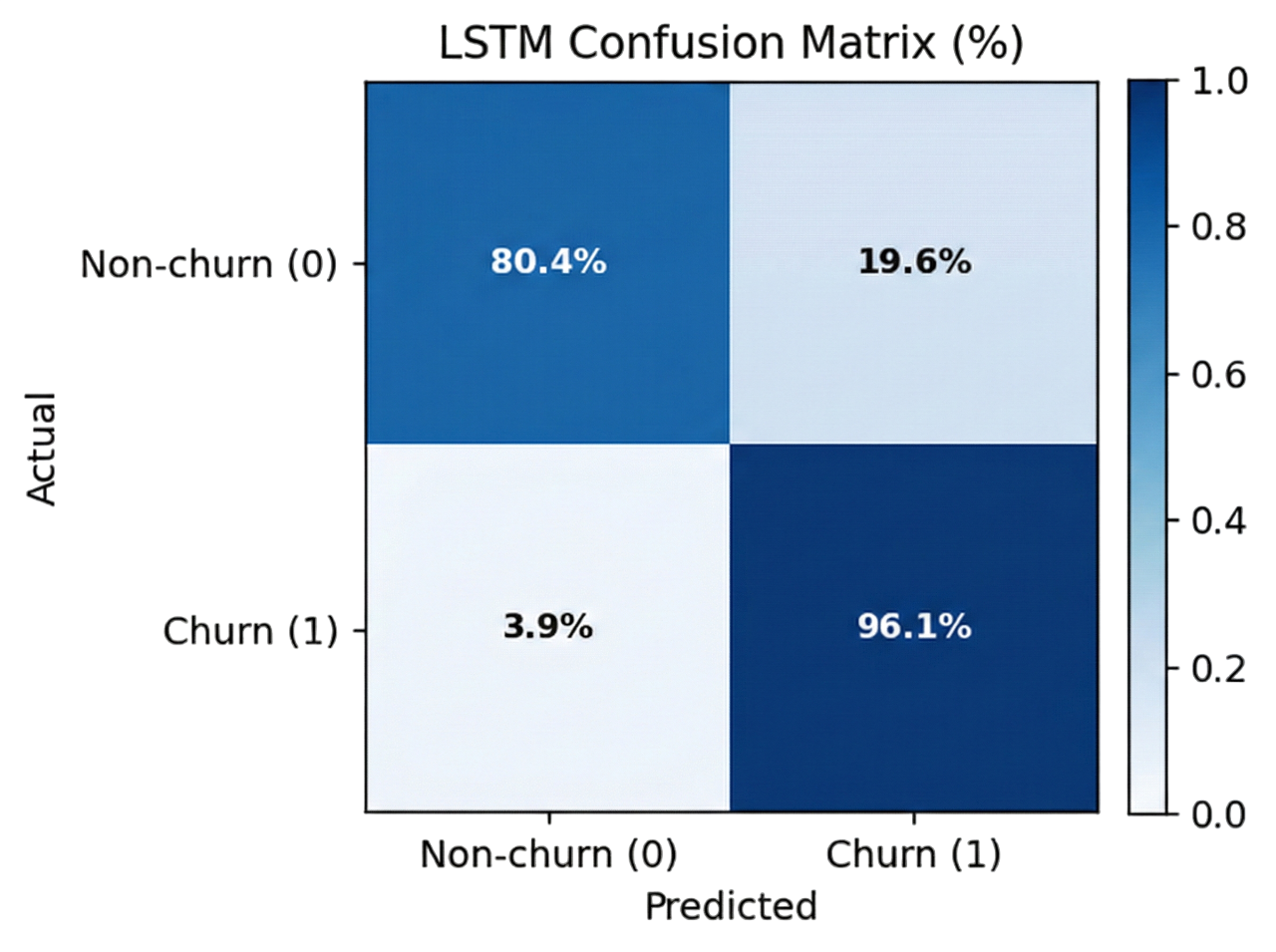}
\caption{Confusion matrices for (a) XGBoost and (b) LSTM models evaluated on the rolling-window test dataset.}
\label{fig:confusion_matrices}
\end{figure*}

\subsubsection{XGBoost Performance}
Table~\ref{tab:model_performance} reports the predictive performance of the XGBoost model on the rolling-window data. The model demonstrates strong and balanced classification behavior across both training and test sets, with consistent precision, recall, and F1-score values. This indicates that the feature-based approach effectively captures short-term behavioral patterns without pronounced overfitting.

The discriminative capability of the model is further illustrated by the ROC curve in Fig.~\ref{fig:roc_comparison}(a). The corresponding confusion matrix shown in Fig.~\ref{fig:confusion_matrices}(a) indicates high correct classification rates for both churn and non-churn users, reflecting a balanced trade-off between sensitivity and specificity.

\subsubsection{LSTM Performance}
The performance of the LSTM model on the rolling-window data is summarized in Table~\ref{tab:model_performance}. Compared to XGBoost, the LSTM achieves higher overall test accuracy and exhibits notably stronger sensitivity toward churn users, as reflected by its recall. This behavior suggests that the sequence-based model is particularly effective at capturing temporal disengagement patterns.

\begin{figure}[t]
\centering
\includegraphics[width=\columnwidth]{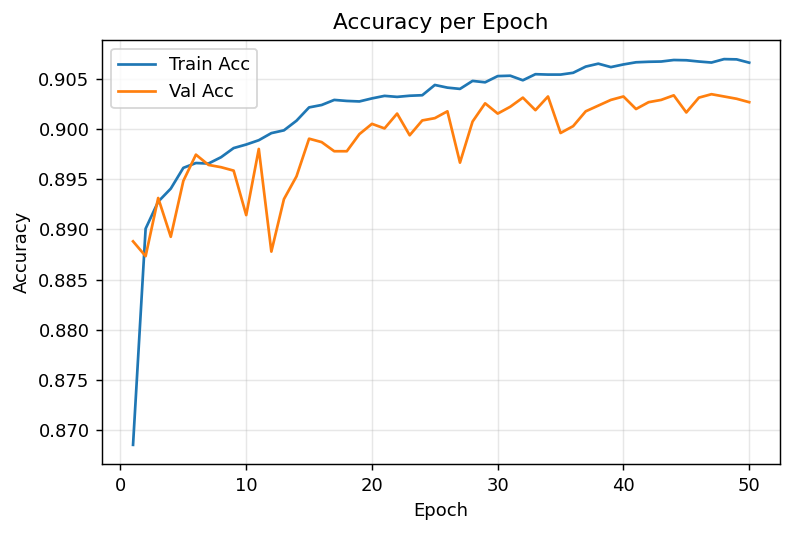}
\caption{Training and validation accuracy of the LSTM model across epochs.}
\label{fig:lstm_acc}
\end{figure}

\begin{figure}[t]
\centering
\includegraphics[width=\columnwidth]{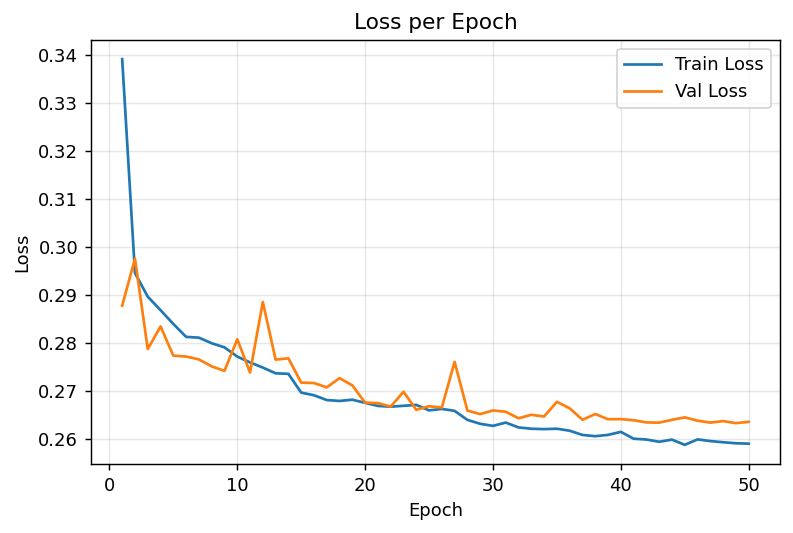}
\caption{Training and validation loss of the LSTM model across epochs.}
\label{fig:lstm_loss}
\end{figure}

Fig.~\ref{fig:roc_comparison}(b) illustrates the ROC curve for the LSTM model, confirming competitive discriminative performance. The confusion matrix in Fig.~\ref{fig:confusion_matrices}(b) shows a higher correct detection rate for churn users, accompanied by an increased rate of false positives for non-churn users. The per-epoch training and validation accuracy trends shown in Fig.~\ref{fig:lstm_acc}, together with the corresponding per-epoch training and validation loss trajectories presented in Fig.~\ref{fig:lstm_loss}, indicate stable convergence behavior and the absence of severe overfitting during training.

\subsection{Generalizability to Future Booking Behavior}

To evaluate temporal robustness under realistic deployment conditions, the trained models are applied to a temporally held-out dataset representing future customer behavior from June--July 2025. This dataset is processed using the same preprocessing, rolling-window construction, feature engineering, and churn labeling pipeline as the historical data, and the models are evaluated without retraining. The resulting generalizability performance is summarized in Table~\ref{tab:generalizability}.

The XGBoost model maintains an accuracy of 83.14\% and a ROC-AUC of 0.911 on the future dataset, indicating strong temporal stability under distributional shift. The LSTM model achieves an accuracy of 81.18\% and exhibits a high recall of 91.80\% for churn users, although its ROC-AUC declines to 0.830. These results indicate that while both models generalize meaningfully to future booking behavior, the feature-based model demonstrates greater overall stability, whereas the sequence-based model preserves stronger sensitivity to churn in an out-of-time evaluation setting.

\begin{table*}[t]
\centering
\caption{Generalizability performance of XGBoost and LSTM models on unseen June 2025 data}
\label{tab:generalizability}
\begin{tabular}{lccccc}
\hline
\textbf{Model} & \textbf{Accuracy} & \textbf{Precision} & \textbf{Recall} & \textbf{F1-score} & \textbf{ROC-AUC} \\
\hline
XGBoost & 83.14\% & 83.29\% & 76.13\% & 79.55\% & 0.911 \\
LSTM    & 81.18\% & 83.91\% & 91.80\% & 87.68\% & 0.830 \\
\hline
\end{tabular}
\end{table*}

\begin{table*}[t]
\centering
\caption{Top factors contributing to churn based on XGBoost gain importance (cumulative threshold = 0.95)}
\label{tab:gain_importance}
\begin{tabular}{lcc}
\hline
\textbf{Feature} & \textbf{Gain} & \textbf{Cumulative Gain} \\
\hline
trend\_total\_bookings & 48.26 & 0.123 \\
weekend\_success\_rate & 35.38 & 0.214 \\
total\_bookings & 27.51 & 0.284 \\
days\_since\_last\_booking & 23.46 & 0.411 \\
payment\_type\_most\_frequent & 21.16 & 0.465 \\
completed\_bookings & 15.86 & 0.505 \\
only\_one\_booking & 12.68 & 0.538 \\
last\_booking\_month & 10.10 & 0.564 \\
trend\_completion\_rate & 9.76 & 0.589 \\
avg\_wallet\_usage & 9.48 & 0.613 \\
first\_booking\_month & 8.93 & 0.636 \\
\hline
\end{tabular}
\end{table*}

\begin{figure*}[t]
\centering
\includegraphics[width=0.47\textwidth]{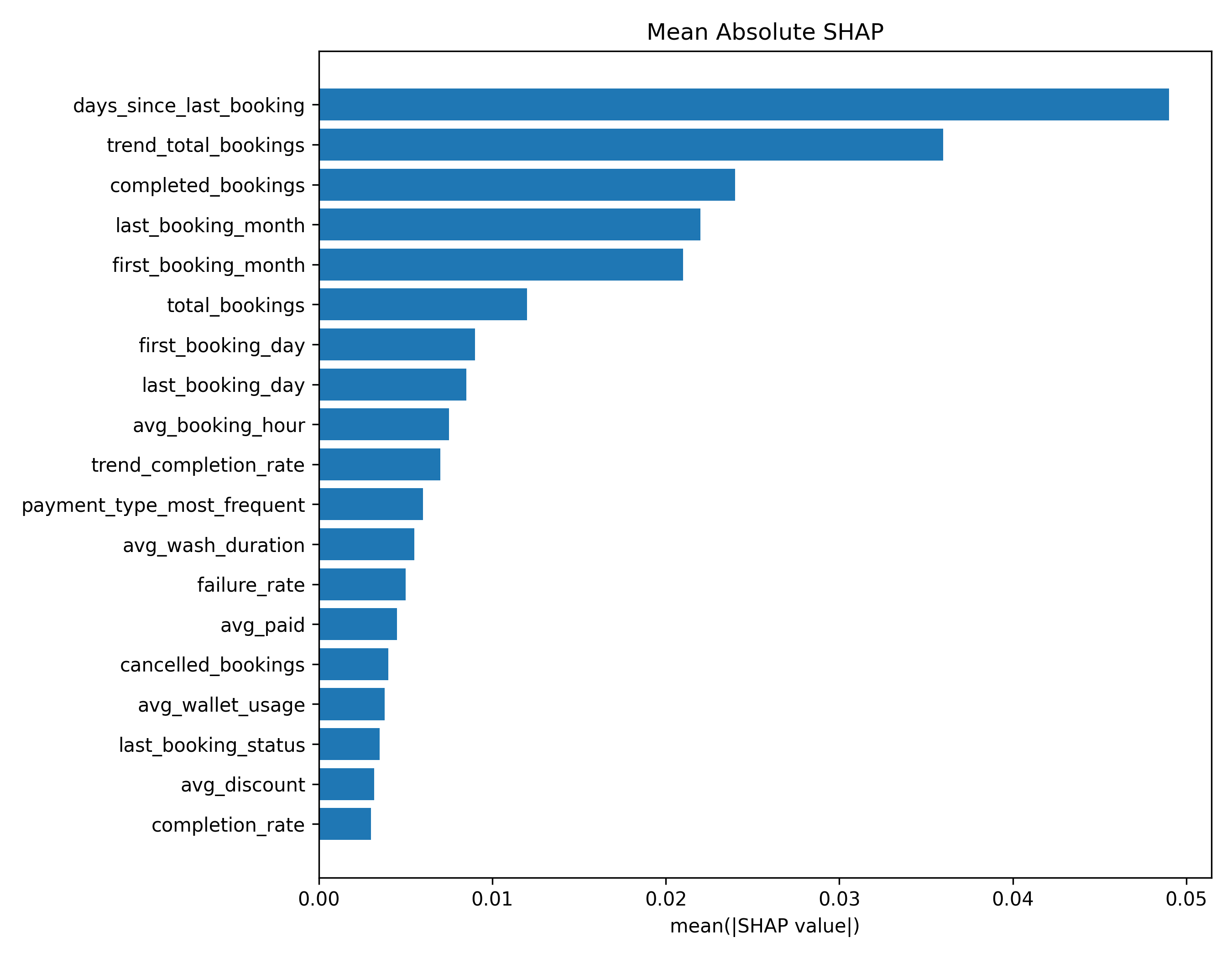}
\hfill
\includegraphics[width=0.42\textwidth]{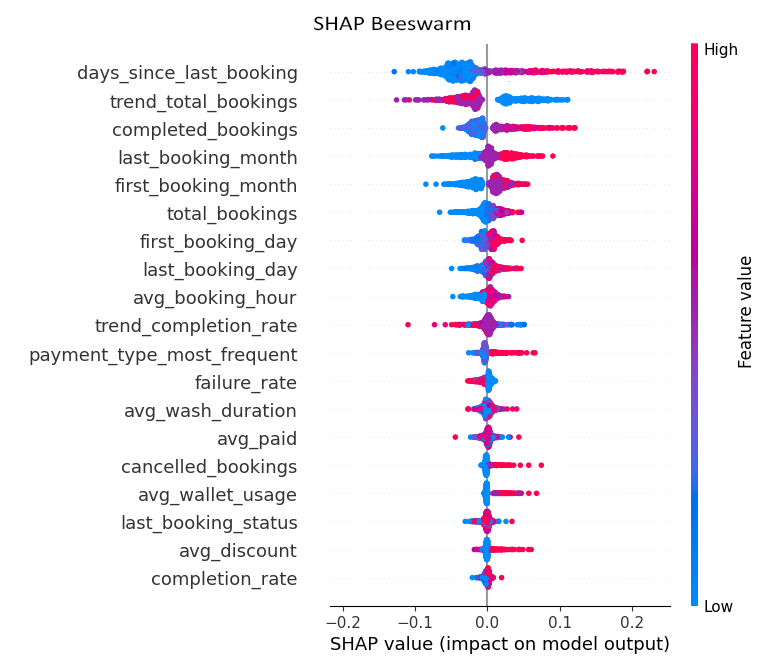}
\caption{SHAP-based interpretation of the XGBoost model: (a) Mean absolute SHAP values indicating global feature importance, and (b) SHAP beeswarm plot illustrating the distribution and direction of feature effects on churn prediction.}
\label{fig:shap_analysis}
\end{figure*}

\subsection{Factors Contributing to Customer Churn}
To identify the behavioral drivers underlying churn predictions, gain-based feature importance and SHAP analyses were conducted for the XGBoost model. The top contributing features based on cumulative gain importance are reported in Table~\ref{tab:gain_importance}. Features related to short-term engagement dynamics, including \textit{trend total bookings}, \textit{weekend success rate}, \textit{total bookings}, and \textit{days since last booking}, emerged as the dominant global predictors.

SHAP-based explanations provide further insight into the directionality and consistency of these effects. The SHAP mean absolute importance values and beeswarm plots, shown in Figure~\ref{fig:shap_analysis}, reveal that declining booking trends, prolonged inactivity, and sparse engagement significantly increase churn probability. Conversely, recent consistent activity and higher booking success rates are associated with reduced churn risk. The alignment between gain-based and SHAP-based explanations confirms the robustness and interpretability of the identified churn drivers.

\section{Discussion}
\label{sec:discussion}
This study set out to examine how short-term behavioral windowing and temporal evaluation influence churn prediction performance in non-contractual, pay-per-use service environments. The results demonstrate that explicitly modeling recent customer behavior through rolling windows enables both accurate and operationally meaningful churn prediction, while preserving interpretability and temporal robustness.

\subsection{Effectiveness of Sliding Behavioral Windows}

A key contribution of this work lies in the adoption of a daily sliding-window formulation, where repeated behavioral instances are generated for the same customer using a fixed observation horizon followed by an explicit churn window. Unlike static snapshot approaches that summarize a customer’s entire history into a single record, the proposed framework captures evolving engagement patterns and supports continuous churn risk assessment.

Compared to prior studies that employ fixed or coarse-grained temporal aggregation, such as monthly or quarterly panels \cite{Mena2024}, the finer-grained sliding windows used in this study enable earlier detection of disengagement signals. This design choice is particularly important in consumer-facing on-demand services contexts, where customer activity is highly volatile and engagement levels can change rapidly. The empirical findings suggest that relatively short observation windows, when enriched with trend-based and recency features, are sufficient to capture churn-relevant dynamics without relying on long historical spans.

\subsection{Feature-Based versus Sequence-Based Learning under Rolling Windows}

The comparative evaluation of XGBoost and LSTM models highlights the complementary strengths of feature-based and sequence-based learning within the same temporal framework. The feature-based XGBoost model demonstrates strong stability and balanced classification behavior across rolling-window test data and future unseen bookings, suggesting that engineered behavioral summaries can effectively capture non-linear interactions among short-term engagement indicators.

In contrast, the LSTM model exhibits higher sensitivity to churn users, reflecting its ability to model sequential dependencies across consecutive behavioral windows. This finding aligns with prior work that emphasizes the value of sequence modeling for early churn detection \cite{Ahlstrand2025}. However, the observed trade-off between recall and overall stability also underscores the importance of careful temporal validation when deploying deep learning models in real-world settings. Unlike several prior studies that employ advanced learning models on cross-sectional or implicitly aggregated behavioral features without explicit temporal prediction horizons \cite{Vo2021,Lalwani2022}, the present work demonstrates that sequence models derive their primary benefit when paired with a principled windowing strategy.

\subsection{Temporal Generalizability and Deployment Robustness}

An important aspect of this study is the evaluation of model performance on a temporally held-out dataset representing future customer behavior. Many existing churn prediction studies report strong in-sample or random-split performance but do not assess robustness under temporal shift \cite{DeCaigny2020,Geiler2022}. By contrast, the results presented here show that both models maintain meaningful predictive capability when applied to future booking data processed through the same pipeline.

The feature-based XGBoost model exhibits greater temporal stability, while the LSTM model retains strong recall for churn users despite some degradation in overall discrimination. These findings are consistent with observations reported in recent window-based churn studies that explicitly account for temporal drift \cite{Bugajev2025}. From a decision support perspective, this highlights an important trade-off between stability and sensitivity that practitioners must consider when selecting models for deployment.

\subsection{Interpretability of Churn Drivers under Rolling Evaluation}

Beyond predictive performance, the proposed framework emphasizes interpretability through gain-based feature importance and SHAP analysis. The convergence of both methods on a consistent set of dominant churn drivers, such as recency of activity, short-term booking trends, and engagement density, reinforces the behavioral validity of the learned models. Importantly, these explanations are derived from repeated behavioral instances rather than static customer profiles, enabling insight into how churn risk evolves over time.

This temporal interpretability distinguishes the present work from prior explainable churn models that operate on cross-sectional data \cite{DeCaigny2024}. By linking feature importance to rolling windows, the framework supports actionable decision-making, allowing service providers to identify not only who is at risk of churn, but also when and why that risk increases.

\subsection{Implications for Decision Support Systems}

From a Decision Support Systems perspective, the findings suggest that effective churn prediction requires more than high predictive accuracy on static benchmarks. The integration of sliding behavioral windows, temporal validation, and interpretable modeling enables continuous risk monitoring and supports proactive intervention strategies. Compared to more complex architectures such as transformers \cite{Ahlstrand2025}, the proposed framework demonstrates that competitive performance can be achieved with relatively lightweight models when temporal framing is carefully designed.

Overall, this study provides evidence that short-term behavioral modeling, combined with rolling evaluation and explainable analytics, offers a practical and robust foundation for churn-oriented decision support in mobility and other on-demand service platforms.

\section{Conclusion}\label{sec:conclusion}

This study proposes a temporally explicit churn prediction framework for non-contractual, pay-per-use service environments, formulating churn prediction as a rolling, instance-level task that enables repeated churn risk assessment as customer behavior evolves. By explicitly defining behavioral observation windows and future churn evaluation horizons, the framework aligns churn modeling with realistic deployment conditions and overcomes limitations of static and cross-sectional approaches. Empirical results demonstrate strong and stable predictive performance, with the feature-based XGBoost model exhibiting balanced behavior and greater robustness under temporal shift. In contrast, the sequence-based LSTM model shows higher sensitivity to churn users by capturing sequential disengagement patterns. The integration of gain-based feature importance and SHAP analysis further supports interpretable identification of behaviorally meaningful churn drivers, reinforcing the framework’s suitability for decision-oriented use. Overall, the findings highlight that carefully designed temporal framing, rather than model complexity alone, is central to achieving deployment-ready and interpretable churn prediction; future work may explore adaptive windowing strategies, drift-aware learning mechanisms, and extensions to cost-sensitive evaluation and additional non-contractual service domains.

% \section*{Declaration of competing interest}

% None.

% \section*{Funding Statement}
% This research did not receive any specific grant from funding agencies in the public, commercial, or not-for-profit sectors.

% \section*{Data Availability Statement}
% The data used in this study were provided by an industry partner and contain commercially sensitive and confidential information. Due to contractual and privacy restrictions, the dataset cannot be made publicly available. Access to the data is therefore restricted, and the authors are unable to share the raw data used in this research.

% =========================
% REFERENCES
% =========================

\bibliography{references}

% \bio{figs/Jawad.png}
% Muhammad Jawad Mufti is a Master's student in Computer Science at King Fahd University of Petroleum and Minerals, Saudi Arabia. His research focuses on temporal user behavior modeling, personalization, recommender systems, and human–computer interaction. He is interested in developing data-driven decision support methods for understanding and improving user experiences.
% \endbio

% \vskip3pc

% \bio{figs/DrOmar.jpg}
% Dr. Omar Hammad is an Assistant Professor at King Fahd University of Petroleum and Minerals. He earned a PhD in Computer Science from the University of Colorado Boulder, with a specialization in Human-Computer Interaction. His research spans software usability, interaction design, user behavior analysis, and social computing research.
% \endbio

% \bio{figs/DrHaitham.jpg}
% Dr. Haitham H. Saleh is an Associate Professor of Industrial and Systems Engineering at KFUPM and affiliated with the Interdisciplinary Research Center for Smart Mobility and Logistics. He earned a BS/MS from KFUPM, an MS from UIUC, and a PhD from Purdue (2018). His interests include optimization, AI, healthcare, and decision support.
% \endbio

% \bio{figs/DrMuqaddas.jpg}
% Dr. Muqaddas Gull earned a BS in Software Engineering (2013) and an MS in Computer Science (2015) from the University of Sargodha, Pakistan, and a PhD in Computer Science from NUST (2024). She is a researcher at the SDAIA–KFUPM AI Center, where she focuses on zero-shot learning, computer vision, and deep learning.
% \endbio

\end{document}